%% file: neurips_2026.tex
\documentclass{article}

\PassOptionsToPackage{numbers, compress}{natbib}
\usepackage[main, final]{neurips_2026}

\usepackage[utf8]{inputenc} 
\usepackage[T1]{fontenc}    
\usepackage{hyperref}       
\usepackage{url}            
\usepackage{booktabs}       
\usepackage{amsfonts}       
\usepackage{nicefrac}       
\usepackage{microtype}      
\usepackage{xcolor}         
\usepackage{amsmath}
\usepackage{caption}
\usepackage{booktabs}      
\usepackage{multirow}
\usepackage{graphicx}
\usepackage{float}
\usepackage{pifont}
\usepackage{url}            
\usepackage{booktabs}       
\usepackage{amsfonts}       
\usepackage{nicefrac}       
\usepackage{microtype}      
\usepackage{amsmath, amssymb, amsthm}
\usepackage{colortbl}
\usepackage{pifont} 
\usepackage{multirow}
\usepackage{multicol}
\usepackage{makecell}
\usepackage{graphicx}
\usepackage{diagbox}
\usepackage{subcaption}
\usepackage[utf8]{inputenc}
\usepackage{svg}
\usepackage{comment}

\usepackage{float}
\usepackage{enumitem}
\usepackage{caption}
\usepackage{graphicx}
\usepackage{caption}
\usepackage{booktabs}
\usepackage{graphicx}    
\usepackage{caption}     
\usepackage{subcaption}  
\usepackage{algorithm}
\usepackage{algpseudocode}
\usepackage{amsmath}

\algrenewcommand\algorithmicindent{1.0em}

\title{Dynamic Video Generation: Shaping Video Generation Across Time and Space}

\author{%
  David S.~Hippocampus\thanks{Use footnote for providing further information
    about author (webpage, alternative address)---\emph{not} for acknowledging
    funding agencies.} \\
  Department of Computer Science\\
  Cranberry-Lemon University\\
  Pittsburgh, PA 15213 \\
  \texttt{hippo@cs.cranberry-lemon.edu} \\
}

\author{ \bf
Shikang Zheng\textsuperscript{1,2*}\thanks{Equal contribution. \dag Corresponding author.},
Jingkai Huang\textsuperscript{2*},
Jiacheng Liu\textsuperscript{1},
Guantao Chen\textsuperscript{1},\\ \bf
Lixuan He\textsuperscript{3},
Yuqi Lin\textsuperscript{1},
Peiliang Cai\textsuperscript{1},
Linfeng Zhang\textsuperscript{1\dag}
\\[0.3em]
\textsuperscript{1}Shanghai Jiao Tong University,
\textsuperscript{2}South China University of Technology,
\textsuperscript{3}Tsinghua University\\
}

\begin{document}

\maketitle

\vspace{-3mm}
\begin{abstract}
Diffusion models have achieved impressive performance in video generation, but their iterative denoising process remains computationally expensive due to the large number of tokens processed at each timestep. Recently, progressive resolution sampling has emerged as a promising acceleration approach by reducing latent resolution in early stages. However, scaling this idea to video generation remains challenging, as the additional temporal dimension introduces diverse spatio-temporal demands across different videos, and compressing only a single dimension often leads to limited acceleration or degraded quality. Therefore, we propose \textbf{DVG}, a \textbf{D}ynamic \textbf{V}ideo \textbf{G}eneration framework that jointly allocates computation across time and space, automatically selecting content-aware acceleration strategies without manual tuning or retraining. \textbf{DVG} achieves near-lossless acceleration across models and tasks, reaching up to \textbf{7$\times$} speedup on HunyuanVideo and HunyuanVideo-1.5, and \textbf{18$\times$} when combined with distillation, demonstrating its potential as a key component in today’s large-scale efficient video generation systems. Our code is available at: \emph{\textcolor[HTML]{57402F}{https://github.com/DarrenZheng303/DVG}}.
\end{abstract}

\begin{figure}[htp]
  \vspace{-9mm}
  \centering
  \includegraphics[trim=2 150 170 120, clip,width=1\linewidth]{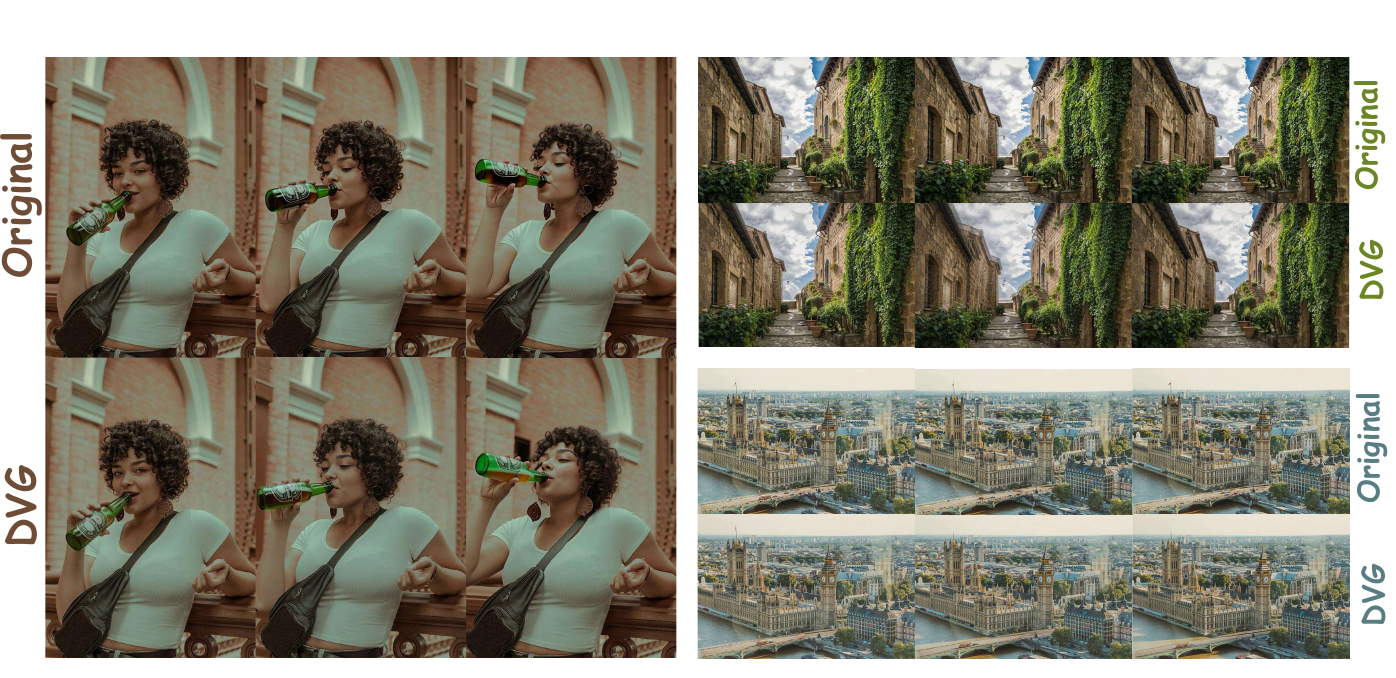}
  \vspace{-6mm}
  \caption{Videos generated on \textbf{HunyuanVideo-1.5} using \textbf{DVG} with distillation at \textbf{18$\times$} speedup.}

  \label{fig:poster}
\end{figure}

\vspace{-1mm}

\input{sec/1_intro.tex}

\input{sec/2_related.tex}

\input{sec/3_method.tex}
\input{sec/4_exp.tex}

\input{sec/5_discussion.tex}

\input{sec/6_conclusion.tex}

\section*{Acknowledgment}

This paper was supported by CCF-Tencent Rhino-Bird Funds.

\bibliography{neurips_2026}
\bibliographystyle{abbrvnat}






\clearpage

\appendix

\section{Appendix}

\subsection{Detailed Experiment Settings}
\subsubsection{Model Configurations.}
\paragraph{HunyuanVideo T2V.}
For the HunyuanVideo text-to-video backbone, we evaluate our method on $720\times1280$ resolution, 125 frames, with 50 denoising steps and a fixed random seed of 42. The sampler follows the default configuration, using an embedded CFG scale of 6.0, a flow shift of 7.0, and reverse-flow sampling.
\paragraph{HunyuanVideo-1.5 T2V.}
For the HunyuanVideo-1.5 text-to-video backbone, videos are generated from the VBench text prompt at the recommended $480\text{p}$ resolution with a $16{:}9$ aspect ratio and 121 frames. We use 50 denoising steps, a fixed random seed of 42, \texttt{bf16} precision, and a flow shift of 7.0.

\paragraph{HunyuanVideo-1.5 I2V.}
For the HunyuanVideo-1.5 image-to-video backbone, we use the CFG-distilled \texttt{480p\_i2v\_distilled} checkpoint. Videos are generated from the reference image and text prompt at the recommended $480\text{p}$ resolution with a $16{:}9$ aspect ratio and 121 frames. We use 50 denoising steps, a fixed random seed of 42, \texttt{bf16} precision, and a flow shift of 7.0.

\paragraph{Wan2.2 T2V.}
For the Wan2.2 text-to-video backbone, we use the MoE-based \texttt{Wan2.2-T2V-A14B} checkpoint. We evaluate on VBench using the default 480p configuration, with a resolution of $832\times480$, 81 frames, and 40 denoising steps. We use a fixed random seed of 42 and follow the repository default mixed-precision setup, where both the T5 encoder and diffusion model use \texttt{bf16} weights. Sampling follows the default Wan2.2 T2V configuration with the UniPC sampler, a sample shift of 12.0, and a two-stage classifier-free guidance scale of $(3.0, 4.0)$, with the expert boundary set to 0.875. For large-scale evaluation, inference is distributed with FSDP and DeepSpeed Ulysses using 2 GPUs per worker.

\subsubsection{DVG Inference Settings.}
For our method, we use a Three-stage dynamic inference schedule in our experiments. The first stage starts from a fixed coarse configuration, where the spatial resolution is simply set to 0.5 of the target resolution and the temporal length is set to 13 latent frames, corresponding to approximately 42\% of the full 31-latent-frame video. The final stage restores the full target resolution and frame count. The intermediate stages are determined automatically according to our action matching algorithm.

\subsubsection{Benchmarks.}
\paragraph{VBench.}
For text-to-video evaluation, we use the official VBench~\citep{VBench} protocol with the full prompt set from \texttt{VBench\_full\_info.json}, which contains 946 prompt entries. We evaluate all VBench dimensions used in our setup, including subject consistency, imaging quality, background consistency, motion smoothness, overall consistency, human action, multiple objects, spatial relationship, object class, color, aesthetic quality, appearance style, temporal flickering, scene, temporal style, and dynamic degree. We report the normalized VBench scores, including the aggregated Quality Score, Semantic Score, and Total Score.

\paragraph{VBench-I2V.}
For image-to-video evaluation, we use the official VBench-I2V protocol with the full prompt set from \texttt{vbench2\_i2v\_full\_info.json}, which contains 1118 image-prompt pairs. We evaluate all VBench-I2V dimensions used in our setup, including i2v subject, i2v background, camera motion, subject consistency, background consistency, aesthetic quality, imaging quality, temporal flickering, motion smoothness, and dynamic degree. We report the normalized VBench-I2V scores together with the aggregated Quality Score, I2V Score, and Total Score.

For the aggregated scores, the Quality Score measures intrinsic visual-temporal quality, while the Semantic Score in VBench measures text-video alignment. In VBench-I2V, the I2V Score measures image-conditioned generation faithfulness, including the consistency of the generated video with the input image and the camera-motion instruction. The Total Score is the weighted aggregate of the corresponding normalized scores. Among individual dimensions, imaging quality captures frame-level artifacts such as blur, noise, and over-exposure, while dynamic degree measures whether the generated video contains sufficient motion.

\subsection{Timing Breakdown.}
We further report the timing breakdown of our method on HunyuanVideo-1.5 I2V at the $480\text{p}$ setting with 121 frames, using 50 denoising steps and a density budget of 0.5, measured on an NVIDIA RTX PRO 6000 GPU. The total generation time is $99.04$s, where the preparation before entering the denoising loop takes $0.36$s. The early sketch stage takes $4.18$s. The action decision module, including clean latent prediction, demand calculation, and best-action scoring, takes only $0.01$s in total, while latent upsampling takes $0.003$s. The spatial-temporally compressed stages take $31.20$s in total. The refinement stage, which runs at the full target resolution and frame rate, takes $53.77$s. The final VAE decoding step takes $9.51$s. This breakdown shows that the additional overhead introduced by action selection and upsampling is negligible compared with the main denoising computation.

\subsection{Limitations.}
Although DVG demonstrates strong acceleration across multiple video generation models and tasks, it still has several limitations. First, our current implementation relies on a discrete set of spatio-temporal compression actions. While this design makes the acceleration ratio controllable and avoids additional training, a richer or continuous learnable action space may further improve the quality-efficiency trade-off for more diverse video contents. Second, DVG estimates spatial and temporal demands from lightweight latent-space cues, which are efficient and generally effective, but may not fully capture all perceptual factors in highly complex scenes, such as subtle object interactions or fine-grained camera motion. Finally, our experiments focus on representative open-source video diffusion models and standard benchmarks; extending the evaluation to more architectures, longer videos, and additional real-world deployment settings would further validate the generality of the framework. We leave these directions for future work.

\newpage
\input{checklist.tex}

\end{document}

%% file: sec/1_intro.tex
\begin{figure}[htp]

  \centering
  \includegraphics[trim=30 280 150 150, clip,width=1\linewidth]{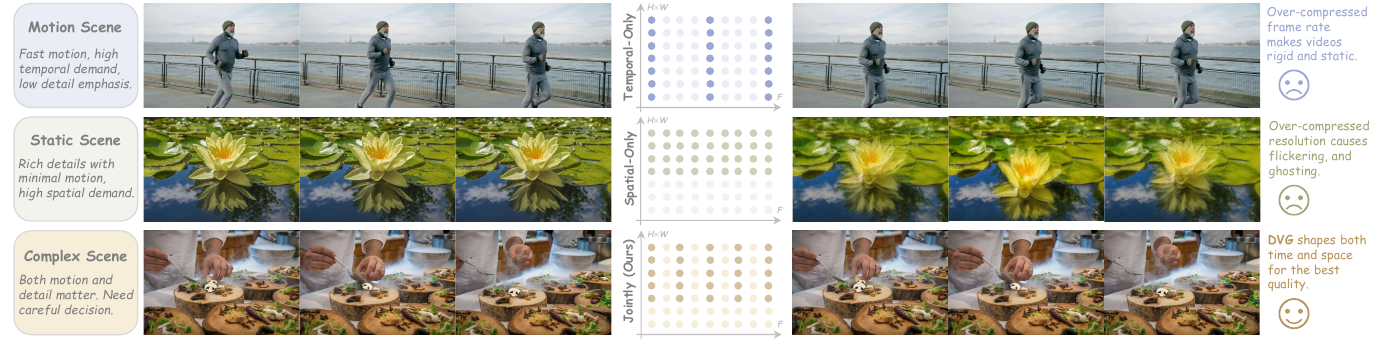}

  \caption{\textbf{Different videos require different spatio-temporal compression strategies.} Compressing only a single dimension often causes motion rigidity or visual artifacts.}

  \vspace{-3mm}

  \label{fig:intro}
\end{figure}

\section{Introduction}

Diffusion models have recently achieved remarkable performance in video generation, delivering state-of-the-art fidelity and diversity. However, their strong generation capability comes with substantial inference cost. Due to the iterative nature of diffusion sampling, each sample typically requires dozens of forward passes through a large transformer backbone, making real-time generation and deployment on resource-constrained devices highly challenging. This computational bottleneck has motivated extensive research on efficient diffusion inference.

Existing acceleration methods can be broadly divided into two directions. One line of work reduces the number of sampling steps using advanced numerical solvers~\citep{lu2022dpm, zheng2023dpmsolvervF}, distillation~\citep{salimans2022progressive}, or consistency training~\citep{song2023consistency}. Another line of work reduces the computation of each denoising step by optimizing model execution, such as sparse attention~\citep{child2019generatinglongsequencessparse, zaheer2021bigbirdtransformerslonger}, token merging~\citep{bolya2023tokenmergingfaststable,bolya2023tokenmergingvitfaster,lu2025tomatokenmergeattention}. For efficient large-scale video generation models, recent progress has been largely driven by step distillation~\citep{zheng2026largescalediffusiondistillation,zou2026discaacceleratingvideodiffusion}, which reduces sampling to only a few denoising steps. As the step count decreases, the per-step cost of processing massive spatio-temporal tokens becomes the dominant bottleneck. Recently, progressive resolution sampling~\citep{zheng2026sketchfrescoefficientdiffusion,tian2025trainingfreediffusionaccelerationbottleneck,zhang2025trainingfreeefficientvideogeneration} has emerged as a promising technique to accelerate image generation: it performs early denoising at lower latent resolutions to reduce the computational cost of spatial tokens~\citep{jeong2026trainingfreemixedresolutionlatentupsampling,tian2025trainingfreediffusionaccelerationbottleneck}. However, scaling this idea to video generation introduces new challenges.

Unlike images, video generation introduces an additional temporal dimension. Video prompts naturally exhibit diverse spatio-temporal demands. Some prompts describe static scenes with rich textures, where preserving spatial details is more important than using a dense frame rate. Others describe fast actions or complex camera motions, where temporal continuity is more critical than fine details. Moreover, achieving a substantial acceleration ratio by compressing only a single dimension often comes at a high cost: reducing spatial resolution causes flickering and unstable details, while reducing frame rate weakens motion dynamics, as shown in Fig.~\ref{fig:intro}. Therefore, an efficient video generation system should not only use a fixed acceleration schedule on one dimension, but adapt its compression strategy according to the video content on both spatial and temporal dimensions.

This nature further motivates a different view of video acceleration: instead of manually choosing a fixed compression ratio as an isolated heuristic, we formulate it as a budgeted spatio-temporal resource allocation problem. A video model spends computation over both space and time: spatial tokens determine how much visual structure and detail can be refined, while temporal tokens determine how densely motion is represented. Under a fixed compute budget, allocating more spatial tokens to one dimension necessarily leaves fewer affordable tokens for the other. Therefore, effective video acceleration requires an optimization mechanism that automatically balances spatial and temporal computation according to the video content.

Motivated by these observations, we propose \textbf{DVG}, a \textbf{D}ynamic \textbf{V}ideo \textbf{G}eneration framework that automatically selects the acceleration strategy based on the video content and jointly compress the latent resolution and frame during denoising. It first performs a few low-cost denoising steps to obtain a coarse latent sketch, from which it estimates spatial detail and motion demands directly in latent space within \textbf{0.01s}, avoiding expensive VAE decoding. Given a target compute budget, DVG automatically selects the best spatio-temporal compression ratio without manual tuning, and finally upsamples the latent back to the original settings. Extensive experiments across multiple models and tasks show that DVG achieves \textbf{4$\times$} speedup on Wan2.2, \textbf{7$\times$} speedup on HunyuanVideo and HunyuanVideo-1.5 while preserving generation quality. When combined with distillation, DVG further reaches \textbf{18$\times$} speedup, as illustrated in Fig.~\ref{fig:poster}. In summary, our main contributions are:

\begin{itemize}[leftmargin=10pt,topsep=0pt]
\item \textbf{Spatio-temporal View of Video Acceleration.}
We revisit video diffusion acceleration from a joint space-time  perspective. Since compressing only a single dimension leads to limited speedup or noticeable quality degradation, we propose DVG, a dynamic video generation framework that jointly optimizes spatial and temporal compression.
    
\item \textbf{Content-Aware Allocation.}
We formulate spatio-temporal allocation as a content-aware optimization problem. DVG automatically estimates spatial and temporal demands from latent sketches, and selects the best compression strategy under a target compute budget without manual tuning.
    
\item \textbf{Outstanding Performance.}
We evaluate DVG on various models and tasks, including the dense model HunyuanVideo, the MoE-based model Wan2.2, and the efficient video generation system HunyuanVideo-1.5. DVG consistently achieves strong acceleration while preserving quality.
\end{itemize}

%% file: sec/2_related.tex
\section{Related Work}

Diffusion models~\citep{sohl2015deep,ho2020DDPM} have demonstrated remarkable capabilities in various generation tasks. While early approaches predominantly relied on U-Net architectures~\citep{ronneberger2015unet}, their scalability limitations were later addressed by Diffusion Transformers~\citep{peebles2023dit}, which have emerged as the dominant backbone for large-scale generative models. In particular, many recent state-of-the-art video generation models adopt DiT-based architectures, enabling high-resolution and long-duration video synthesis~\citep{chen2023pixartalpha,chen2024pixartsigma,opensora,yang2025cogvideox}. Despite these advances, the iterative sampling process remains a fundamental limitation of diffusion models, introducing substantial computational overhead during inference. This has driven two primary research directions: reducing the number of sampling steps and lowering the per-step computational cost. Beyond efficiency, a key challenge is to preserve generative fidelity and stability, especially under aggressive acceleration.

\subsection{Sampling Timestep Reduction}

DDIM~\citep{songDDIM} introduced deterministic few-step sampling that preserves perceptual quality. Higher-order ODE solvers (DPM-Solver and variants)~\citep{lu2022dpm,lu2022dpm++,zheng2023dpmsolvervF} further improve the accuracy–efficiency trade-off via multi-step and multi-stage discretizations with controlled local truncation error. Rectified Flow~\citep{refitiedflow} shortens transport paths, while distillation methods~\citep{salimans2022progressive} compress long sampling trajectories into compact generators. Consistency models~\citep{song2023consistency} also enable few-step generation by learning a direct noise-to-clean mapping. Feature caching~\citep{liu2025reusingforecastingacceleratingdiffusion,zheng2025forecastcalibratefeaturecaching,zheng2025letfeaturesdecidesolvers} methods reuse intermediate features across timesteps to reduce redundant computation, effectively lowering the number of actual computed timesteps. Recently, step-distillation~\citep{zheng2026largescalediffusiondistillation,zou2026discaacceleratingvideodiffusion} has emerged as the dominant acceleration paradigm in practice, directly reducing the number of inference steps while maintaining strong generation quality. Its simplicity and effectiveness have led to widespread adoption in large-scale video generation.

\subsection{Per-Step Computation Reduction}

Pruning~\citep{structural_pruning_diffusion,zhu2024dipgo}, quantization~\citep{10377259,shang2023post,kim2025ditto}, and token reduction~\citep{bolya2023tomesd,kim2024tofu,zhang2024tokenpruningcachingbetter,zhang2025sito,cheng2025catpruningclusterawaretoken} reduce computational cost with relatively low runtime overhead. In parallel, sparse attention methods, including fixed-window, strided, and adaptive patterns~\citep{beltagy2020longformerlongdocumenttransformer, zaheer2021bigbirdtransformerslonger, child2019generatinglongsequencessparse}, aim to alleviate the quadratic complexity of self-attention by restricting token interactions, though their practical speedups are often limited. To achieve more substantial acceleration, recent efforts have focused on reducing spatial resolution during generation. Cascade diffusion frameworks~\citep{luo2023videofusiondecomposeddiffusionmodels, ho2021cascadeddiffusionmodelshigh, li2021srdiffsingleimagesuperresolution, saharia2021imagesuperresolutioniterativerefinement} generate low-resolution outputs and progressively refining them via learned upsampling, but often require retraining and auxiliary modules. Progressive resolution sampling methods~\citep{tian2025trainingfreediffusionaccelerationbottleneck,zheng2026sketchfrescoefficientdiffusion,jeong2026trainingfreemixedresolutionlatentupsampling}, adapt resolution across sampling stages, offering improved flexibility but relying on carefully designed schedules and hyperparameters, which makes it hard to scale them to large video models with complex spatiotemporal dependencies.

%% file: sec/3_method.tex
\begin{figure}[htp]

  \centering
  \includegraphics[trim=60 130 65 130, clip,width=1\linewidth]{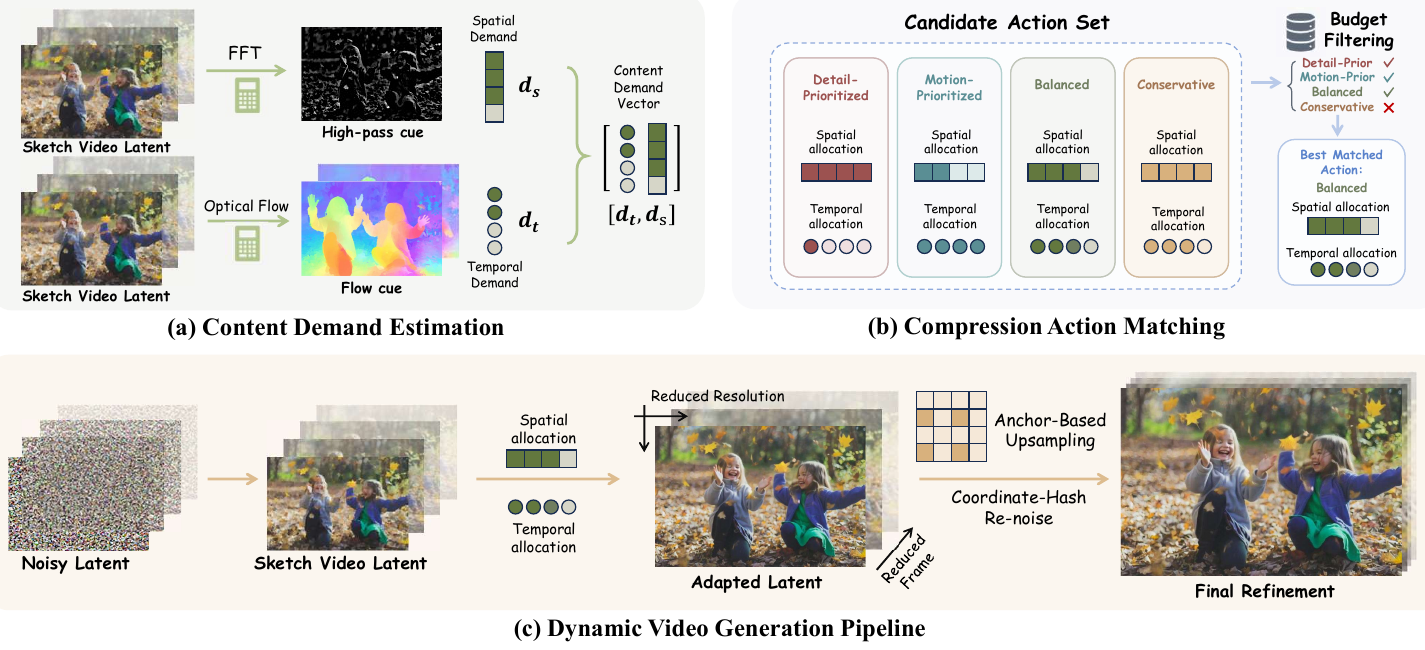}

  \caption{\textbf{DVG Framework.} DVG first predicts a coarse latent video sketch, then analyzes its spatial and temporal demands directly in latent space. Under a target compute budget, DVG selects the best compression action for denoising, then restores the latent to the original setting for refinement.}

  \label{fig:intro}
\end{figure}

\section{Method}

\subsection{Preliminary}

\noindent\textbf{Diffusion Models.}
Diffusion models generate samples by simulating a forward–reverse stochastic process. In the forward process, a clean sample \(x_0\) is progressively corrupted by Gaussian noise:
\begin{equation}
    x_t = \sqrt{\alpha_t} x_0 + \sqrt{1 - \alpha_t} \, \epsilon_t, \quad \epsilon_t \sim \mathcal{N}(0, \mathbf{I}),
\end{equation}
where \(\alpha_t\) defines a predefined noise schedule. The reverse process learns to recover \(x_0\) by predicting the noise component \(\epsilon_\theta(x_t, t)\), leading to the update:
\begin{equation}
    x_{t-1} = \frac{1}{\sqrt{\alpha_t}} \left( x_t - \frac{1 - \alpha_t}{\sqrt{1 - \bar{\alpha}_t}} \, \epsilon_\theta(x_t, t) \right) + \sigma_t \, \epsilon,
\end{equation}
where \(\bar{\alpha}_t = \prod_{s=1}^t \alpha_s\), and \(\sigma_t\) controls the stochasticity. In video diffusion models, the latent variable \(x_t\) represents a spatiotemporal feature tensor, and each denoising step involves high-dimensional computation across both spatial and temporal dimensions, leading to substantial inference cost.

\noindent\textbf{Progressive Resolution Sampling.} Progressive resolution methods accelerate diffusion by performing early denoising on a lower-resolution latent grid, where fine details are less important. Given a latent feature map \(\mathbf{x}_t \in \mathbb{R}^{H \times W \times C}\), the process is:
\[
\hat{\mathbf{x}}_t = \mathcal{U}_s\big(f(\mathcal{D}_s(\mathbf{x}_t))\big),
\]
where \(\mathcal{D}_s\) and \(\mathcal{U}_s\) are downsampling and upsampling operators, and \(f(\cdot)\) denotes denoising on the coarse grid. By reducing the spatial tokens, these methods reduced intermediate computation.

\subsection{Overview of DVG}

DVG is a content-adaptive acceleration framework for video diffusion transformers that jointly adjusts spatial resolution and temporal frame-rate during denoising. The key observation is that different videos require different computational allocation: videos with rich textures benefit more from high spatial resolution, while videos with large motions require denser temporal sampling. Instead of using a fixed resolution-frame schedule for all prompts, DVG dynamically selects an efficient spatio-temporal denoising action according to an early preview of the generated video.

DVG first performs a few denoising steps at low spatial resolution and low frame-rate to obtain a noisy but informative intermediate latent. It then predicts a coarse clean latent in one step as an early video sketch. Based on this sketch, DVG estimates the spatial detail demand and temporal motion demand in latent space. Under a user-specified computational budget, DVG searches a pre-defined action list and selects the action whose spatial and temporal allocation best matches the estimated content demand. After denoising the intermediate stage with the selected action, the latent is upsampled to full spatial resolution and full frame-rate for final refinement.

\subsection{Sketching Early Perception of Video Content}

At the beginning of denoising, the latent \(X_0\) is dominated by Gaussian noise and contains little semantic information, making early compression decisions unreliable. To obtain a lightweight preview of the video content, DVG first performs a few denoising steps under a low resolution and low frame-rate setting, producing a sketch latent \(X_k\) that already captures coarse motion and structure. Instead of running the remaining denoising process for schedule estimation, we directly extrapolate a rough clean latent from the current flow prediction. Specifically, given the model prediction \(v_\theta(X_k, k)\) at step \(k\), we estimate:
\[
\tilde{X}
=
X_k
+
\alpha_k \, v_\theta(X_k, k),
\]
where \(\alpha_k\) is a timestep-dependent scaling factor controlling the extrapolation strength. This produces a coarse video latent that is sufficient for estimating motion intensity and spatial detail complexity, while avoiding expensive full-resolution denoising.

\subsection{Inferring Content Demands in Latent Space}

Given the predicted latent sketch \(\tilde{X} \in \mathbb{R}^{C \times F \times H \times W}\), DVG estimates two complementary content demands: spatial demand and temporal demand. The spatial demand reflects the amount of fine-grained visual detail, while the temporal demand measures motion intensity and temporal variation.

\noindent\textbf{Spatial demand.}
DVG estimates spatial complexity with Fast Fourier Transform. It applies a high-pass operator \(\mathcal{H}(\cdot)\) to each latent frame and compute the normalized high-frequency energy:
\begin{equation}
d_s
=
\operatorname{Norm}
\left(
\frac{1}{F}
\sum_{t=1}^{F}
\left\|
\mathcal{H}(\tilde{X}_t)
\right\|_2
\right),
\label{eq:spatial-demand}
\end{equation}
where \(\tilde{X}_t\) denotes the \(t\)-th latent frame. Larger \(d_s\) values indicate richer textures and more detailed scene structures, suggesting that more computation should be allocated to spatial resolution.

\noindent\textbf{Temporal demand.}
To estimate motion complexity, DVG computes optical flow on the latent sketch. Let \(\mathcal{F}_{t \rightarrow t+1}\) denote the optical flow between adjacent frames. The temporal demand is defined as:
\begin{equation}
d_t
=
\operatorname{Norm}
\left(
\frac{1}{F-1}
\sum_{t=1}^{F-1}
\left\|
\mathcal{F}_{t \rightarrow t+1}
\right\|_2
\right).
\label{eq:temporal-demand}
\end{equation}
Larger \(d_t\) values indicate stronger motion and more frequent temporal changes, implying that higher temporal resolution should be preserved. Then, DVG normalized the two demand into weights:
\begin{equation}
m_s
=
\frac{
\exp(\alpha d_s)
}{
\exp(\alpha d_s) + \exp(\alpha d_t)
},
\qquad
m_t = 1 - m_s,
\label{eq:demand-ratio}
\end{equation}
where \(\alpha\) controls the allocation sharpness. Larger \(m_s\) favors richer spatial tokens, while larger \(m_t\) favors more temporal tokens.

\subsection{Constructing the Space of Spatio-Temporal Actions}

To make the acceleration ratio controllable, DVG pre-defines an action list $\mathcal{A}$. Each action $a \in \mathcal{A}$ specifies the spatial and temporal scales used across denoising stages:
\begin{equation}
a =
\left\{
(r^1_s, r^1_t),
(r^2_s, r^2_t),
\cdots,
(r^L_s, r^L_t)
\right\},
\label{eq:action}
\end{equation}
where $r^\ell_s$ and $r^\ell_t$ denote the spatial compression ratio and temporal compression ratio at stage $\ell$, at a grid of 0.05. For example, an action may use $(0.5, 0.7)$ in the intermediate stage, meaning that denoising is performed with $50\%$ original latent resolution and $70\%$ frame-rate.

For an action $a$, its computational density is estimated as:
\begin{equation}
\rho(a)
=
\frac{
\sum_{\ell=1}^{L}
H_\ell W_\ell F_\ell \cdot N_\ell
}{
H W F \cdot N
},
\label{eq:density}
\end{equation}
where $H_\ell = r^\ell_s H$, $W_\ell = r^\ell_s W$, $F_\ell = r^\ell_t F$, and $N_\ell$ is the number of denoising steps assigned to stage $\ell$. Here, $H$, $W$, $F$, and $N$ are the full-resolution height, width, frame count, and total denoising steps. The density $\rho(a)$ approximates the relative computation cost compared with full-resolution full-frame-rate denoising. Given a target budget density $D$ and tolerance $\epsilon$, DVG retains only actions whose computational density satisfies:
\begin{equation}
\mathcal{A}_B
=
\left\{
a \in \mathcal{A}
\mid
|\rho(a)-D| \leq \epsilon
\right\}.
\label{eq:budget-filter}
\end{equation}
This budget filtering step ensures that the selected action satisfies the desired acceleration level before content-aware action matching is applied.

\subsection{Matching Actions to Content Demands}

After obtaining the budget-feasible action set $\mathcal{A}_B$, DVG selects the candidate action according to how well it matches the spatial and temporal demands. For each action $a$, we define its effective spatial and temporal allocation as:
\begin{equation}
g_s(a) = \frac{1}{L}\sum_{\ell=1}^{L} r^\ell_s,
\qquad
g_t(a) = \frac{1}{L}\sum_{\ell=1}^{L} r^\ell_t,
\label{eq:gain}
\end{equation}
where \(L\) denotes the number of denoising stages. Intuitively, $g_s(a)$ measures how much spatial resolution the action preserves, while $g_t(a)$ measures how much temporal frame-rate it preserves.

The final action is then selected by solving the following demand matching objective:
\begin{equation}
a^\star
=
\arg\max_{a \in \mathcal{A}_B}
\left[
-\lambda
\left(
g_s(a) - m_s
\right)^2
-
(1-\lambda)
\left(
g_t(a) - m_t
\right)^2
\right],
\label{eq:action-selection}
\end{equation}
where $\lambda$ balances spatial and temporal matching. This formulation encourages DVG to allocate more spatial tokens to texture-rich videos and more frame to motion-intensive videos, while still satisfying the overall target computational budget.

\subsection{Flexible Spatio-Temporal Latent Reshaping}

When transitioning to a different resolution and frame-rate, DVG first resizes the current latent to the target spatio-temporal grid and then re-noises it at the next timestep. For each axis with source length \(K\) and target length \(N\), we assign source position \(i\) to an anchor index:
\begin{equation}
a_i=\operatorname{round}\left(\frac{i(N-1)}{K-1}\right).
\end{equation}
Each target position \(p\) is then interpolated from neighboring anchors:
\begin{equation}
y(p)=(1-\beta)x(i)+\beta x(i+1),
\qquad
\beta=\frac{p-a_i}{a_{i+1}-a_i}.
\end{equation}
Applying this operation along temporal, height, and width axes produces the resized latent \(\bar{X}\). Next, DVG generates coordinate-aligned Gaussian noise directly on the target grid. For each coordinate \((t,h,w,c)\) and random seed \(s\), a deterministic hash generates uniform variables, which are converted into Gaussian noise using the Box--Muller transform:
\begin{equation}
\epsilon_{\mathrm{coord}}(t,h,w,c)
=
\sqrt{-2\log u_1}\cos(2\pi u_2),
\qquad
u_1,u_2=\operatorname{Hash}(t,h,w,c,s).
\end{equation}
The latent at the next timestep is then obtained through scheduler re-noising:
\begin{equation}
X_{\tau}^{\mathrm{next}}
=
\bar{X}
+
\sigma(\tau)\epsilon_{\mathrm{coord}}.
\end{equation}

Unlike prior progressive sampling methods that are usually restricted to fixed \(2\times\) spatial or temporal scaling, our anchor-based resizing and coordinate-hash re-noising support arbitrary target resolutions and frame-rates. This enables DVG to flexibly reshape the latent space and construct a much richer compression action space for content-aware allocation. Finally, the refine denoising stage is always performed at full resolution and frame-rate to further capture visual details and temporal consistency.

%% file: sec/4_exp.tex
\section{Experiments}

\noindent\textbf{Model Configurations.}
We evaluate DVG on both text-to-video (T2V) and image-to-video (I2V) tasks across models with different architectures. For T2V, we conduct experiments on dense video model {HunyuanVideo}~\citep{sun_hunyuanvideo_2024} at \(720\text{p}\) with 125 frames. To further validate generality, we also evaluated the MoE-based model {Wan2.2-14B}~\citep{wan2025} at \(480\text{p}\) with 81 frames, as well as the recent efficient video generation system {HunyuanVideo-1.5}~\citep{wu2025hunyuanvideo15technicalreport} in the recommended setting \(480\text{p}\), 121-frames. For I2V generation, we evaluate DVG on {HunyuanVideo-1.5} using the same 121-frame configuration. These experiments allow us to evaluate DVG across different model architectures and examine its compatibility with modern -oriented designs.

\noindent\textbf{Evaluation and Metrics.}
T2V results are measured using VBench~\citep{VBench}, which contains 946 prompts and provides a comprehensive multi-dimensional evaluation of video quality. We report the key metrics: Quality, Semantic, and Total Scores. 
I2V evaluation follows VBench-I2V~\citep{VBench} with 1118 prompts, where I2V Scores and Total scores are reported. 
In addition, we include PSNR, SSIM, and LPIPS to assess its reconstruction fidelity. 
\emph{Please refer to the Appendix for detailed settings.}

\input{tab/hyvideo}

\subsection{Results on Text-to-Video Generation}

\begin{figure}[htp]
  \centering
  \includegraphics[trim=160 110 180 100, clip,width=1\linewidth]{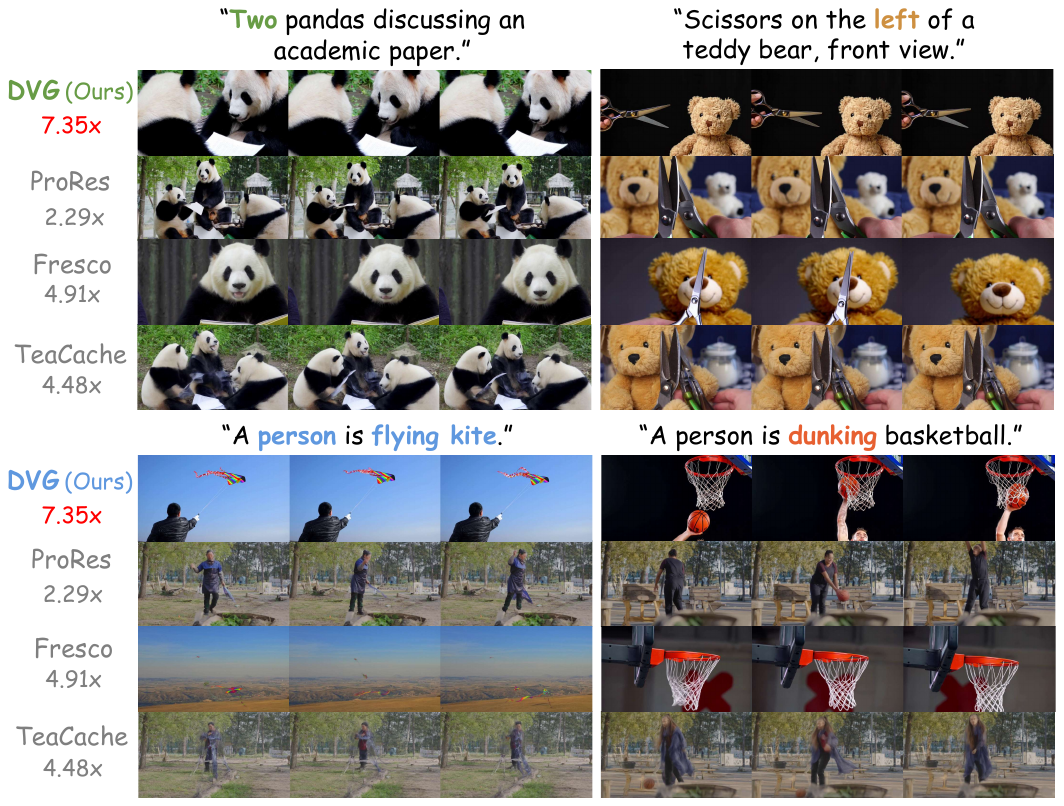}

  \caption{\textbf{Visualization of different acceleration methods on HunyuanVideo.} DVG achieves better semantic alignment and visual quality, whereas prior methods degrade under high acceleration.}

  \label{fig:hy1.0}
\end{figure}

As summarized in Table~\ref{table:hyvideo}, under the standard 50-step setting and density 50\%, DVG delivers a \textbf{3.20$\times$} speedup while maintaining a high total score of \textbf{80.45}, outperforming all single-dimension compression and caching methods with even better efficiency. Under more aggressive compression, DVG further reaches \textbf{7.35$\times$} speedup with a total score of \textbf{80.14} using only 25 sampling steps, surpassing all baselines and demonstrating strong robustness with fewer-step generation. Visualization in Fig.~\ref{fig:hy1.0} further demonstrates DVG’s superior efficiency while preserving quality.

Results on Wan2.2 and HunyuanVideo-1.5 are summarized in Table~\ref{table:hy1.5_t2v}. On the MoE-based Wan2.2, DVG achieves \textbf{3.76$\times$} speedups while maintaining less than 1\% quality drop. On HunyuanVideo-1.5, it consistently outperforms both baseline and SpargeAttention. With density 70\%, it achieves the best total score of \textbf{80.51}. Under stronger acceleration, DVG still maintains a high total score of \textbf{79.77} at \textbf{6.58$\times$} speedup. These results demonstrate that DVG generalizes well across different architectures while providing a favorable efficiency and quality.

\input{tab/hyvideo1.5_t2v}

\subsection{Results on Image-to-Video Generation}

DVG consistently achieves strong performance on image-to-video generation, as shown in Table~\ref{table:hy1.5_i2v}. Under the CFG-distilled setting, it delivers a 3.79$\times$ speedup while achieving \textbf{23.03} PSNR and \textbf{0.101} LPIPS, indicating high similarity to the original video. When applying with R-MeanFlow, DVG further reaches up to \textbf{18.09$\times$} speedup, while still achieving excellent perceptual quality (\textbf{23.92} PSNR, \textbf{0.081} LPIPS) and the highest I2V score. Compared to other step-reduction and single-dimension compression methods, DVG achieves significantly better reconstruction quality while providing the highest additional speedup, demonstrating strong compatibility with distillation, as shown in Fig.~\ref{fig:hy1.5}.

\input{tab/hyvideo1.5_i2v}

\input{tab/ablation}

%% file: tab/hyvideo.tex
\begin{table*}[htb]
\centering
\caption{\textbf{Quantitative comparison of text-to-video generation} on HunyuanVideo.}

\setlength\tabcolsep{5.0pt} 
\small
\resizebox{\textwidth}{!}{
\begin{tabular}{l | c | c  c | c  c  c}
    \toprule
    \multirow{2}{*}{\centering \bf Method} 
    & \multirow{2}{*}{\centering \bf NFE} 
    & \multicolumn{2}{c|}{\bf Acceleration} 
    & \multicolumn{3}{c}{\bf VBench Score (\%)} \\
    
    \cline{3-7}
     &  & {\bf Latency(s) $\downarrow$} & {\bf Speed $\uparrow$} 
     & {\bf Quality $\uparrow$} 
     & {\bf Semantic $\uparrow$} 
     & {\bf Total $\uparrow$} 
     \rule{0pt}{2ex}\\ 
    \midrule

HunyuanVideo~\citep{sun_hunyuanvideo_2024}
    & 50  & 1947.38 & 1.00$\times$ 
    & 82.97 \textcolor{gray!70}{\scriptsize (+0.0\%)} 
    & 71.41 \textcolor{gray!70}{\scriptsize (+0.0\%)} 
    & 80.66 \textcolor{gray!70}{\scriptsize (+0.0\%)} \\ 
\midrule

SpargeAttn~\citep{zhang2025spargeattentionaccuratetrainingfreesparse}
    & 50  & 1094.01 & 1.78$\times$ 
    & 83.26 \textcolor{gray!70}{\scriptsize (+0.3\%)} 
    & 71.24 \textcolor{gray!70}{\scriptsize (-0.2\%)} 
    & 80.35 \textcolor{gray!70}{\scriptsize (-0.4\%)} \\

LIPAR~\citep{menn2026videocompressionmeetsvideo}
    & 50  & 1421.44 & 1.37$\times$ 
    & 81.72 \textcolor{gray!70}{\scriptsize (-1.5\%)} 
    & 69.14 \textcolor{gray!70}{\scriptsize (-3.2\%)} 
    & 79.63 \textcolor{gray!70}{\scriptsize (-1.3\%)} \\

ToMe~\citep{bolya2023tokenmergingfaststable}
    & 50  & 1343.02 & 1.45$\times$ 
    & 80.65 \textcolor{gray!70}{\scriptsize (-2.8\%)} 
    & 67.87 \textcolor{gray!70}{\scriptsize (-4.9\%)} 
    & 78.39 \textcolor{gray!70}{\scriptsize (-2.8\%)} \\

VGDFR~\citep{yuan2025vgdfrdiffusionbasedvideogeneration}
    & 50  & 1209.55 & 1.61$\times$ 
    & 80.45 \textcolor{gray!70}{\scriptsize (-3.0\%)} 
    & 69.45 \textcolor{gray!70}{\scriptsize (-2.7\%)} 
    & 79.03 \textcolor{gray!70}{\scriptsize (-2.0\%)} \\

Jenga-ProRes~\citep{zhang2025trainingfreeefficientvideogeneration}
    & 50  & 1361.38 & 1.43$\times$ 
    & \textbf{83.53} \textcolor{gray!70}{\scriptsize (+0.7\%)} 
    & 70.11 \textcolor{gray!70}{\scriptsize (-1.8\%)} 
    & 80.07 \textcolor{gray!70}{\scriptsize (-0.7\%)} \\

Bottleneck~\citep{tian2025trainingfreediffusionaccelerationbottleneck}
    & 30  & 818.23 & 2.38$\times$ 
    & 81.97 \textcolor{gray!70}{\scriptsize (-1.2\%)} 
    & 69.55 \textcolor{gray!70}{\scriptsize (-2.6\%)} 
    & 79.42 \textcolor{gray!70}{\scriptsize (-1.5\%)} \\
TeaCache~\citep{liu2025timestepembeddingtellsits} $({l}=0.15)$
    & 50 & 861.23 & 2.26$\times$ 
    & 82.09 \textcolor{gray!70}{\scriptsize (-1.1\%)} 
    & 69.17 \textcolor{gray!70}{\scriptsize (-3.1\%)} 
    & 79.51 \textcolor{gray!70}{\scriptsize (-1.4\%)} \\

AdaCache~\citep{kahatapitiya2024adaptivecachingfastervideo}
    & 50 & 743.27 & 2.62$\times$ 
    & 81.78 \textcolor{gray!70}{\scriptsize (-1.4\%)} 
    & 70.12 \textcolor{gray!70}{\scriptsize (-1.8\%)} 
    & 79.58 \textcolor{gray!70}{\scriptsize (-1.3\%)} \\

\rowcolor{gray!20}
\textbf{DVG} $(D=50\%)$   
    & 50  & \bf 608.56  & \bf {3.20$\times$} 
    & 82.10 \textcolor[HTML]{0f98b0}{\scriptsize \textbf{(-1.0\%)}} 
    & \bf 73.83 \textcolor[HTML]{0f98b0}{\scriptsize \textbf{(+3.4\%)}} 
    & \bf 80.45 \textcolor[HTML]{0f98b0}{\scriptsize \textbf{(-0.3\%)}} \\

\midrule

{$22\%${ steps}}  
    & 11  & 427.99 & 4.55$\times$ 
    & 80.52 \textcolor{gray!70}{\scriptsize (-3.0\%)} 
    & 69.01 \textcolor{gray!70}{\scriptsize (-3.4\%)} 
    & 77.74 \textcolor{gray!70}{\scriptsize (-3.6\%)} \\

TeaCache~\citep{liu2025timestepembeddingtellsits} $({l}=0.3)$
    & 50 & 522.09 & 3.73$\times$ 
    & 81.55 \textcolor{gray!70}{\scriptsize (-1.7\%)} 
    & 68.73 \textcolor{gray!70}{\scriptsize (-3.8\%)} 
    & 78.99 \textcolor{gray!70}{\scriptsize (-2.1\%)} \\

Jenga-ProRes~\citep{zhang2025trainingfreeefficientvideogeneration}
    & 10  & 417.18 & 4.66$\times$ 
    & 79.24 \textcolor{gray!70}{\scriptsize (-4.5\%)} 
    & 65.57 \textcolor{gray!70}{\scriptsize (-8.2\%)} 
    & 76.50 \textcolor{gray!70}{\scriptsize (-5.2\%)} \\

Fresco~\citep{zheng2026sketchfrescoefficientdiffusion}
    & 20  & 491.76 & 3.96$\times$ 
    & 81.58 \textcolor{gray!70}{\scriptsize (-1.7\%)} 
    & 66.93 \textcolor{gray!70}{\scriptsize (-6.3\%)} 
    & 78.65 \textcolor{gray!70}{\scriptsize (-2.5\%)} \\

\rowcolor{gray!20}
\textbf{DVG} $(D=40\%)$ 
    & 25  & \bf264.95  & \bf {7.35$\times$} 
    & \bf81.59 \textcolor[HTML]{0f98b0}{\scriptsize \textbf{(-1.7\%)}} 
    & \bf 74.33 \textcolor[HTML]{0f98b0}{\scriptsize \textbf{(+4.1\%)}} 
    & \bf 80.14 \textcolor[HTML]{0f98b0}{\scriptsize \textbf{(-0.6\%)}} \\

\bottomrule
\end{tabular}}
\label{table:hyvideo}
\end{table*}

%% file: tab/hyvideo1.5_t2v.tex
\begin{table*}[htb]
\centering
\caption{\textbf{Quantitative comparison of text-to-video generation} on Wan2.2 and HunyuanVideo-1.5.}

\setlength\tabcolsep{5.0pt} 
\small
\resizebox{\textwidth}{!}{
\begin{tabular}{l | c | c  c | c  c  c}
    \toprule
    \multirow{2}{*}{\centering \bf Method} 
    & \multirow{2}{*}{\centering \bf NFE} 
    & \multicolumn{2}{c|}{\bf Acceleration} 
    & \multicolumn{3}{c}{\bf VBench Score (\%)} \\
    
    \cline{3-7}
     &  & {\bf Latency(s) $\downarrow$} & {\bf Speed $\uparrow$} 
     & {\bf Quality $\uparrow$} 
     & {\bf Semantic $\uparrow$} 
     & {\bf Total $\uparrow$} 
     \rule{0pt}{2ex}\\ 
    \midrule

Wan2.2-14B~\citep{wan2025}
    & 40  & 460.27 & 1.00$\times$ 
    & \textbf{85.21} \textcolor{gray!70}{\scriptsize (+0.0\%)} 
    & \textbf{73.37} \textcolor{gray!70}{\scriptsize (+0.0\%)} 
    & \textbf{82.84} \textcolor{gray!70}{\scriptsize (+0.0\%)} \\ 

\rowcolor{gray!20}
\textbf{DVG} $(D=50\%)$
    & 40  & 200.85 & 2.29$\times$ 
    & 85.02 \textcolor[HTML]{0f98b0}{\scriptsize \bf(-0.2\%)} 
    & 72.54 \textcolor[HTML]{0f98b0}{\scriptsize \bf(-1.1\%)} 
    & 82.53 \textcolor[HTML]{0f98b0}{\scriptsize \bf(-0.4\%)} \\ 

\rowcolor{gray!20}
\textbf{DVG} $(D=50\%)$
    & 25  & \bf 122.31 & \bf 3.76$\times$ 
    & 84.61 \textcolor[HTML]{0f98b0}{\scriptsize \bf(-0.7\%)} 
    & 72.33 \textcolor[HTML]{0f98b0}{\scriptsize \bf(-1.9\%)} 
    & 82.15 \textcolor[HTML]{0f98b0}{\scriptsize \bf(-0.8\%)} \\ 
\midrule

HunyuanVideo-1.5~\citep{wu2025hunyuanvideo15technicalreport}
    & 50  & 285.59 & 1.00$\times$ 
    & 83.19 \textcolor{gray!70}{\scriptsize (+0.0\%)} 
    & 62.05 \textcolor{gray!70}{\scriptsize (+0.0\%)} 
    & 78.96 \textcolor{gray!70}{\scriptsize (+0.0\%)} \\ 

SpargeAttn~\citep{zhang2025spargeattentionaccuratetrainingfreesparse}
    & 50  & 186.00 & 1.53$\times$ 
    & 83.83 \textcolor{gray!70}{\scriptsize (+0.8\%)} 
    & 66.52 \textcolor{gray!70}{\scriptsize (+7.2\%)} 
    & 80.36 \textcolor{gray!70}{\scriptsize (+1.8\%)} \\

\rowcolor{gray!20}
\textbf{DVG} $(D=70\%)$ 
    & 50  & 136.73  & {2.09$\times$} 
    & \bf 84.22 \textcolor[HTML]{0f98b0}{\scriptsize \textbf{(+1.2\%)}} 
    & \bf 65.69  \textcolor[HTML]{0f98b0}{\scriptsize \textbf{(+5.9\%)}} 
    & \bf 80.51 \textcolor[HTML]{0f98b0}{\scriptsize \textbf{(+2.0\%)}} \\

\rowcolor{gray!20}
\textbf{DVG} $(D=50\%)$ 
    & 50  & 91.95  & {3.11$\times$} 
    &  84.05 \textcolor[HTML]{0f98b0}{\scriptsize \textbf{(+1.0\%)}} 
    &  65.83  \textcolor[HTML]{0f98b0}{\scriptsize \textbf{(+6.1\%)}} 
    &  80.41 \textcolor[HTML]{0f98b0}{\scriptsize \textbf{(+1.8\%)}} \\

\rowcolor{gray!20}
\textbf{DVG} $(D=50\%)$ 
    & 25  & \bf 43.42  & \bf {6.58$\times$} 
    &  83.39 \textcolor[HTML]{0f98b0}{\scriptsize \textbf{(+0.2\%)}} 
    &  65.26  \textcolor[HTML]{0f98b0}{\scriptsize \textbf{(+5.2\%)}} 
    &  79.77 \textcolor[HTML]{0f98b0}{\scriptsize \textbf{(+1.0\%)}} \\

\bottomrule
\end{tabular}}
\label{table:hy1.5_t2v}
\end{table*}

%% file: tab/hyvideo1.5_i2v.tex
\begin{table*}[htb]
\centering
\caption{\textbf{Quantitative comparison of image-to-video generation} on HunyuanVideo-1.5.}

\setlength\tabcolsep{5.0pt} 
\small
\resizebox{\textwidth}{!}{
\begin{tabular}{l | c | c  c | c  c | c  c  c}
    \toprule
    \multirow{2}{*}{\centering \bf Method} 
    & \multirow{2}{*}{\centering \bf NFE} 
    & \multicolumn{2}{c|}{\bf Acceleration} 
    & \multicolumn{2}{c|}{\bf VBench Score (\%)}
    & \multicolumn{3}{c}{\bf Perceptual Metrics} \\
    
    \cline{3-9}
     &  & {\bf Latency(s) $\downarrow$} & {\bf Speed $\uparrow$} 
     & {\bf I2V $\uparrow$} 
     & {\bf Total $\uparrow$}
     & {\bf PSNR $\uparrow$}
     & {\bf SSIM $\uparrow$}
     & {\bf LPIPS $\downarrow$}
     \rule{0pt}{2ex}\\ 
    \midrule

HunyuanVideo-1.5~\citep{wu2025hunyuanvideo15technicalreport}
    & 100  & 540.04 & 1.00$\times$ 
    & 94.13 
    & 88.45 
    & - & - & - \\ 
\midrule

CFG Distilled~\citep{zou2026discaacceleratingvideodiffusion}
    & 50  & 278.37 & 1.94$\times$ 
    & 93.56 
    & 87.15 
    & $\infty$ & 1.000 & 0.000 \\ 

\hspace{3mm}+ VGDFR~\citep{yuan2025vgdfrdiffusionbasedvideogeneration}
    & 50  & 176.48 & 3.06$\times$ 
    & 91.75 
    & 85.59 
    & 19.37 & 0.683 & 0.145 \\ 

\hspace{3mm}+ AdaCache~\citep{kahatapitiya2024adaptivecachingfastervideo}
    & 50 & 137.76 & 3.92$\times$ 
    & 91.88 
    & 85.71
    & 19.95 & 0.654 & 0.183 \\

\hspace{3mm}+ Jenga-ProRes~\citep{zhang2025trainingfreeefficientvideogeneration}
    & 50  & 199.28 & 2.71$\times$ 
    & 92.35 
    & 86.38 
    & 19.87 & 0.663 & 0.157 \\

\hspace{3mm}+ Bottleneck~\citep{tian2025trainingfreediffusionaccelerationbottleneck}
    & 30  & 132.69 & 4.07$\times$ 
    & 91.23
    & 86.11 
    & 18.86 & 0.681 & 0.201 \\

\rowcolor{gray!20}
\hspace{3mm}+ \textbf{DVG} $(D=70\%)$   
    & 50  & 142.47   & {3.79$\times$} 
    & \bf 92.75 
    & \bf 86.74 
    &  \bf23.03 &  \bf0.710 &  \bf0.101 \\

\rowcolor{gray!20}
\hspace{3mm}+ \textbf{DVG} $(D=50\%)$   
    & 50  &  \bf99.82  &  \bf{5.41$\times$} 
    & 92.86 
    & 86.47 
    & 22.52 & 0.700 & 0.117 \\

\midrule

R-MeanFlow~\citep{zou2026discaacceleratingvideodiffusion}
    & 12  & 75.23 & 7.18$\times$ 
    & 94.10 
    & 87.34 
    & $\infty$ & 1.000 & 0.000 \\

\hspace{3mm}+ VGDFR~\citep{yuan2025vgdfrdiffusionbasedvideogeneration}
    & 12  & 55.33 & 9.76$\times$ 
    & 91.97 
    & 85.25 
    & 19.63 & 0.727 & 0.189 \\ 
    
\hspace{3mm}+ Fresco~\citep{zheng2026sketchfrescoefficientdiffusion}
    & 12  & 52.48 & 10.29$\times$ 
    & 91.85 
    & 84.88 
    & 18.78 & 0.662 & 0.365 \\

\hspace{3mm}+ AdaCache~\citep{kahatapitiya2024adaptivecachingfastervideo}
    & 12 & 52.02 & 10.38$\times$ 
    & 91.79 
    & 85.21 
    & 19.08 & 0.707 & 0.208 \\

\hspace{3mm}+ TaylorSeer~\citep{liu2025reusingforecastingacceleratingdiffusion}
    & 12 & 46.76 & 11.55$\times$ 
    & 92.03 
    & 85.39 
    & 19.34 & 0.714 & 0.205 \\

\rowcolor{gray!20}
\hspace{3mm}+ \textbf{DVG} $(D=70\%)$ 
    &12  & 39.18  & {13.78$\times$} 
    & \bf 93.60 
    & \bf 86.47 
    & \bf23.92 & \bf0.767 & \bf0.081 \\

\rowcolor{gray!20}
\hspace{3mm}+ \textbf{DVG} $(D=50\%)$ 
    &12  & \bf 29.86  & \bf {18.09$\times$} 
    &  93.51 
    &  86.42 
    & 23.73 & 0.766 & 0.083 \\

\bottomrule
\end{tabular}}

\label{table:hy1.5_i2v}
\end{table*}

%% file: tab/ablation.tex
\begin{table*}[htb]
\centering
\caption{\textbf{Ablation Study} on HunyuanVideo-1.5.}

\setlength\tabcolsep{5.0pt} 
\small
\resizebox{\textwidth}{!}{
\begin{tabular}{l | c | c  c | c  c | c  c  c}
    \toprule
    \multirow{2}{*}{\centering \bf Method} 
    & \multirow{2}{*}{\centering \bf Density}
    & \multicolumn{2}{c|}{\bf Avg. Reduction Ratio}
    & \multicolumn{2}{c|}{\bf Acceleration} 
    & \multicolumn{3}{c}{\bf VBench Score (\%)} \\
    
    \cline{3-4} \cline{5-6} \cline{7-9}
     & 
     & {\bf Spatial $\downarrow$} & {\bf Temporal $\downarrow$}
     & {\bf Latency(s) $\downarrow$} & {\bf Speed $\uparrow$} 
     & {\bf Dynamic $\uparrow$} 
     & {\bf Quality $\uparrow$} 
     & {\bf Total $\uparrow$}
     \rule{0pt}{2ex}\\ 
    \midrule

\textbf{DVG (Ours)} 
    & \textasciitilde50\% & 0.29 & 0.46
    & 91.95  & 3.11$\times$ 
    & \underline{43.75} & {84.05} & \underline{80.41} \\

\hspace{3mm}+ {Temporal Only} 
    & \textasciitilde50\% & 0.00 & 0.71
    & 93.03  & 3.07$\times$ 
    & 35.00 & \textbf{85.13} & 78.98 \\

\hspace{3mm}+ {Spatial Only}  
    & \textasciitilde50\% & 0.49 & 0.00
    & {89.54}  & {3.19}$\times$ 
    & \textbf{47.25} & 82.16 & 76.84 \\

\hspace{3mm}+ {Manual Tuning (Fixed Ratio)}  
    & \textasciitilde50\% & 0.30 & 0.45
    & {88.13}  & {3.24}$\times$ 
    & 41.50 & 83.05 & 78.85 \\

\hspace{3mm}+ {Linear Upsampling}  
    & \textasciitilde50\% & 0.29 & 0.46
    & 90.25  & 3.16$\times$ 
    & 43.00 & 79.92 & 79.11 \\

\hspace{3mm}+ {Random Renoise}  
    & \textasciitilde50\% & 0.29 & 0.46
    &{89.37}  & {3.19}$\times$ 
    & {43.50} & 78.37 & 78.92 \\

\hspace{3mm}+ {Pixel-Space Eval}  
    & \textasciitilde50\% & 0.31 & 0.40
    & 102.19  & 2.79$\times$ 
    & \underline{43.75} & \underline{84.83} & \textbf{80.63} \\

\bottomrule
\end{tabular}}
\label{table:ablation}
\end{table*}

%% file: sec/5_discussion.tex
\section{Discussion}

\noindent\textbf{Ablation Study.}
We conduct our ablation study on HunyuanVideo-1.5. Table~\ref{table:ablation} demonstrates that the effectiveness of DVG comes primarily from its adaptive spatial-temporal acceleration design. Compressing only a single dimension leads to clear drops under the same overall density. Moreover, replacing our upsampling with linear interpolation or using random renoise consistently degrades performance, indicating that our tailored upsampling and renoise designs are essential for enabling flexible spatio-temporal compression while maintaining generation consistency. We further compare DVG with manually tuned fixed compression ratios. Although the density budget is similar, DVG achieves significantly better performance through its content-aware allocation strategy, while manually tuned settings often produce inconsistent visual quality across different videos.

\begin{figure}[htp]
  \centering
  \includegraphics[trim=20 135 45 50, clip,width=1\linewidth]{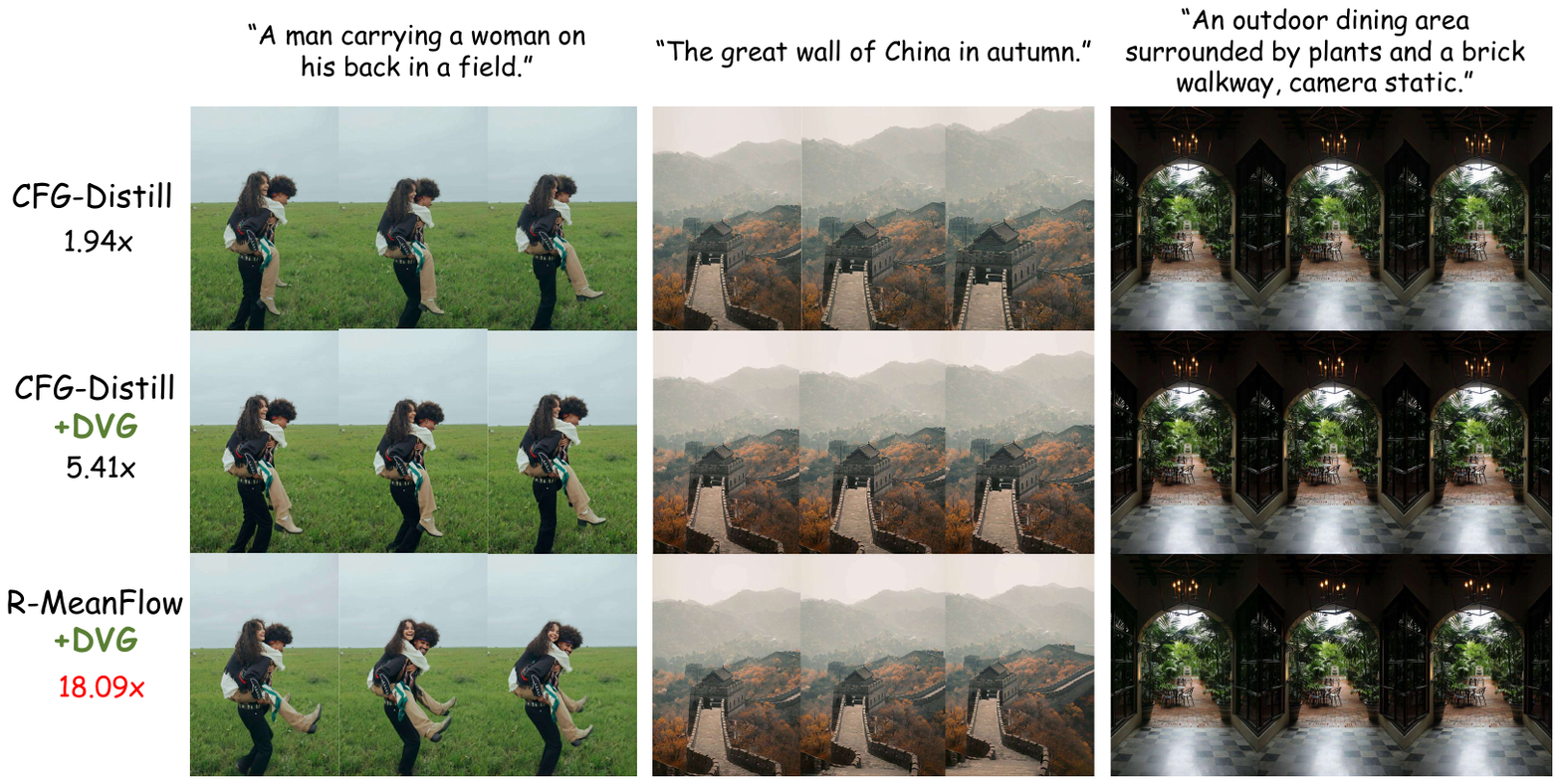}

  \caption{\textbf{Visualization of DVG on HunyuanVideo-1.5.} DVG successfully maintains high-fidelity and motion consistency with the original video, even under distillation, reaching up to 18$\times$ speedup.}

  \label{fig:hy1.5}
\end{figure}

\noindent\textbf{Latent or Pixel Evaluation?}
We compare two strategies for estimating videos' spatial and temporal demand: evaluating directly in latent space versus decoding to pixel space. Across 1118 prompts, the two strategies achieve a Top-1 agreement of 85\% and a Top-2 agreement of 99\%. This suggests that discrepancies arise mainly because of the rich action set, rather than fundamentally different demand estimations. As shown in Table~\ref{table:ablation}, pixel-space evaluation achieves slightly higher VBench scores, but our latent-space evaluation requires only \textbf{0.01s}, because it avoids an additional VAE decoding, thereby significantly reducing end-to-end latency. This advantage becomes increasingly important when combined with other acceleration techniques. For example, in a highly optimized setting (e.g., R-MeanFlow + DVG), generation takes only \textbf{29s}; introducing an extra VAE decode would increase the total latency to \textbf{40s}, resulting in a \textbf{38\%} extra overhead.

%% file: sec/6_conclusion.tex
\section{Conclusion}

In this paper, we revisit efficient video generation from a joint spatial-temporal perspective, showing that accelerating video generation requires balancing both dimensions rather than compressing only one. Based on this observation, we present \textbf{DVG}, a dynamic video generation framework that adaptively allocates computation across space and time under a target compute budget. By estimating content demands directly in latent space and matches the best compression action, DVG achieves strong acceleration while maintaining generation quality across large scale dense models, MoE models, and suits well in modern efficient video generation systems.

%% file: checklist.tex
\clearpage

\section*{NeurIPS Paper Checklist}

\begin{enumerate}

\item {\bf Claims}
    \item[] Question: Do the main claims made in the abstract and introduction accurately reflect the paper's contributions and scope?
    \item[] Answer: \answerYes{} 
    \item[] Justification: The claims match the contributions and experimental scope presented in the paper.

\item {\bf Limitations}
    \item[] Question: Does the paper discuss the limitations of the work performed by the authors?
    \item[] Answer: \answerYes{} 
    \item[] Justification: The paper includes a separate discussion of limitations in appendix.

\item {\bf Theory assumptions and proofs}
    \item[] Question: For each theoretical result, does the paper provide the full set of assumptions and a complete (and correct) proof?
    \item[] Answer: \answerNA{} 
    \item[] Justification: The paper does not present new theoretical results.

    \item {\bf Experimental result reproducibility}
    \item[] Question: Does the paper fully disclose all the information needed to reproduce the main experimental results of the paper to the extent that it affects the main claims and/or conclusions of the paper (regardless of whether the code and data are provided or not)?
    \item[] Answer: \answerYes{} 
    \item[] Justification: The paper provides the experimental setup, baselines, evaluation metrics, and implementation details needed to reproduce the main results.

\item {\bf Open access to data and code}
    \item[] Question: Does the paper provide open access to the data and code, with sufficient instructions to faithfully reproduce the main experimental results, as described in supplemental material?
    \item[] Answer: \answerYes{} 
    \item[] Justification: The code is in supplementary material and will be released on Github.

\item {\bf Experimental setting/details}
    \item[] Question: Does the paper specify all the training and test details (e.g., data splits, hyperparameters, how they were chosen, type of optimizer) necessary to understand the results?
    \item[] Answer: \answerYes{} 
    \item[] Justification: The paper specifies the models, datasets, baselines, metrics, and implementation details used in the experiments.

\item {\bf Experiment statistical significance}
    \item[] Question: Does the paper report error bars suitably and correctly defined or other appropriate information about the statistical significance of the experiments?
    \item[] Answer: \answerNo{} 
    \item[] Justification: The paper reports quantitative experimental results and ablations, but does not currently include error bars, confidence intervals, or statistical significance tests for the main results.

\item {\bf Experiments compute resources}
    \item[] Question: For each experiment, does the paper provide sufficient information on the computer resources (type of compute workers, memory, time of execution) needed to reproduce the experiments?
    \item[] Answer: \answerYes{} 
    \item[] Justification: The paper reports the main compute resources used in our experiments, including RTX PRO 6000 GPUs and the associated vCPU resources.
    
\item {\bf Code of ethics}
    \item[] Question: Does the research conducted in the paper conform, in every respect, with the NeurIPS Code of Ethics \url{https://neurips.cc/public/EthicsGuidelines}?
    \item[] Answer: \answerYes{} 
    \item[] Justification: The research conforms to the NeurIPS Code of Ethics.

\item {\bf Broader impacts}
    \item[] Question: Does the paper discuss both potential positive societal impacts and negative societal impacts of the work performed?
    \item[] Answer: \answerNA{} 
    \item[] Justification: The paper does not include a separate broader impact discussion, as the work focuses on a general inference acceleration method.
    
\item {\bf Safeguards}
    \item[] Question: Does the paper describe safeguards that have been put in place for responsible release of data or models that have a high risk for misuse (e.g., pre-trained language models, image generators, or scraped datasets)?
    \item[] Answer: \answerNA{} 
    \item[] Justification: The paper does not release high-risk models or datasets that require additional safeguards.

\item {\bf Licenses for existing assets}
    \item[] Question: Are the creators or original owners of assets (e.g., code, data, models), used in the paper, properly credited and are the license and terms of use explicitly mentioned and properly respected?
    \item[] Answer: \answerYes{} 
    \item[] Justification: The paper credits the existing datasets, models, codebases, and evaluation tools used in the experiments, and describes their licenses or terms of use.

\item {\bf New assets}
    \item[] Question: Are new assets introduced in the paper well documented and is the documentation provided alongside the assets?
    \item[] Answer: \answerYes{} 
    \item[] Justification: The released code is documented with usage instructions and implementation details.

\item {\bf Crowdsourcing and research with human subjects}
    \item[] Question: For crowdsourcing experiments and research with human subjects, does the paper include the full text of instructions given to participants and screenshots, if applicable, as well as details about compensation (if any)? 
    \item[] Answer: \answerYes{} 
    \item[] Justification: The paper includes a small-scale user study involving voluntary participants who completed anonymous electronic questionnaires for perceptual evaluation of generated videos. The study did not collect sensitive personal information and posed minimal risk to participants. According to our institutional guidelines, this type of anonymous perceptual evaluation does not require formal IRB approval.

\item {\bf Institutional review board (IRB) approvals or equivalent for research with human subjects}
    \item[] Question: Does the paper describe potential risks incurred by study participants, whether such risks were disclosed to the subjects, and whether Institutional Review Board (IRB) approvals (or an equivalent approval/review based on the requirements of your country or institution) were obtained?
    \item[] Answer: \answerNo{} 
    \item[] Justification: Participants were informed of the scope of collected information and its intended use, but the paper does not report IRB approval or an equivalent approval.
    \begin{itemize}
        \item The answer \answerNA{} means that the paper does not involve crowdsourcing nor research with human subjects.
        \item Depending on the country in which research is conducted, IRB approval (or equivalent) may be required for any human subjects research. If you obtained IRB approval, you should clearly state this in the paper. 
        \item We recognize that the procedures for this may vary significantly between institutions and locations, and we expect authors to adhere to the NeurIPS Code of Ethics and the guidelines for their institution. 
        \item For initial submissions, do not include any information that would break anonymity (if applicable), such as the institution conducting the review.
    \end{itemize}

\item {\bf Declaration of LLM usage}
    \item[] Question: Does the paper describe the usage of LLMs if it is an important, original, or non-standard component of the core methods in this research? Note that if the LLM is used only for writing, editing, or formatting purposes and does \emph{not} impact the core methodology, scientific rigor, or originality of the research, declaration is not required.
    \item[] Answer: \answerNA{} 
    \item[] Justification: The core method does not involve LLMs as an important, original, or non-standard component.

\end{enumerate}